\documentclass{article}

\usepackage{microtype}
\usepackage{graphicx}
\usepackage{subfigure}
\usepackage{booktabs} 
\usepackage{multirow}
\usepackage{hyperref}

\usepackage[accepted]{icml2025}
\usepackage{diagbox}

\usepackage{amsmath,amsfonts,bm}

\def\eqref#1{(\ref{#1})}

\def\1{\bm{1}}

\DeclareMathAlphabet{\mathsfit}{\encodingdefault}{\sfdefault}{m}{sl}
\SetMathAlphabet{\mathsfit}{bold}{\encodingdefault}{\sfdefault}{bx}{n}

\DeclareMathOperator*{\argmax}{arg\,max}
\DeclareMathOperator*{\argmin}{arg\,min}

\usepackage{amsmath}
\usepackage{amssymb}
\usepackage{mathtools}
\usepackage{amsthm}
\definecolor{red}{RGB}{247,77,77}
\definecolor{green}{RGB}{14,152,111}
\definecolor{purple}{RGB}{121,108,173}
\definecolor{orange}{RGB}{214,88,19}

\usepackage[capitalize,noabbrev]{cleveref}

\theoremstyle{plain}

\theoremstyle{definition}

\theoremstyle{remark}

\usepackage[textsize=tiny]{todonotes}

\begin{document}

\twocolumn[
\icmltitle{FedAPT: Federated Adversarial Prompt Tuning for Vision-Language Models}

\icmlsetsymbol{equal}{*}

\begin{icmlauthorlist}
\icmlauthor{Kun Zhai}{fudan}
\icmlauthor{Siheng Chen}{jiaoda}
\icmlauthor{Xingjun Ma}{fudan}
\icmlauthor{Yu-Gang Jiang}{fudan}
\end{icmlauthorlist}

\icmlaffiliation{fudan}{Fudan University, Shanghai, China}
\icmlaffiliation{jiaoda}{Shanghai Jiao Tong University, Shanghai, China}

\icmlcorrespondingauthor{Xingjun Ma}{xingjunma@fudan.edu.cn}

\icmlkeywords{Machine Learning, ICML}

\vskip 0.3in
]

\printAffiliationsAndNotice{\icmlEqualContribution}




\begin{abstract}
Federated Prompt Tuning (FPT) is an efficient method for cross-client collaborative fine-tuning of large Vision-Language Models (VLMs). However, models tuned using FPT are vulnerable to adversarial attacks, leading to misclassification in downstream tasks. In this work, we introduce Federated Adversarial Prompt Tuning (\textbf{FedAPT}), a novel method designed to enhance the adversarial robustness of FPT.
We identify a key issue in FedAPT under non-independent and identically distributed (non-IID) settings: a \textit{class information gap} between clients and the global model. Clients rely solely on limited local label information to generate adversarial samples for training, while the global model must defend against adversarial attacks from global labels. To address this issue, we propose a \textbf{class-aware prompt generator} that generates visual prompts from text prompts. This generator is guided by a \emph{Global Label Embedding} (serving as a ``beacon") which encodes cross-client label information to create more globally-aligned visual prompts.
Additionally, we propose a \textbf{cross-layer generator sharing} strategy  to enhance prompt coupling across different layers of the model, further boosting adversarial robustness.
Extensive experiments on multiple image classification datasets demonstrate the superiority of FedAPT in improving adversarial robustness, outperforming existing methods by a large margin. FedAPT also exhibits exceptional generalization in cross-domain and cross-dataset scenarios, indicating its effectiveness in real-world applications.
\end{abstract}

\section{Introduction}\label{sec:Intro}
Federated Prompt Tuning (FPT) has become a crucial approach for privacy-preserving and efficient fine-tuning of large Vision-Language Models (VLMs) like Contrastive Language Image Pretraining (CLIP) \cite{radford2021learning}. By optimizing a small set of additional parameters, FPT significantly reduces computational and communication overhead, making it particularly suitable for privacy-sensitive, distributed environments such as healthcare and biometrics. Despite its advantages, FPT research primarily focuses on two key directions: personalization and domain generalization, while largely overlooking the vulnerability of these methods to adversarial attacks, one major threat to model safety \cite{szegedy2013intriguing, shah2021adversarial}. This work aims to enhance the robustness of FPT against adversarial attacks. 

In the current literature, adversarial training is widely recognized as an effective strategy for enhancing the adversarial robustness of deep learning models \cite{zhang2023delving}. Building on this foundation, researchers have explored integrating prompt tuning with adversarial training to improve the adversarial robustness of tuned VLMs. For instance, AVP \cite{chen2023visual} applies visual prompt tuning to strengthen the adversarial resilience of tuned CLIP, while APT \cite{li2024one} and AdvPT \cite{zhang2025adversarial} utilize text prompts to defend against adversarial attacks on images. Additionally, FAP \cite{zhou2024few}, a bimodal defense method, introduces a multimodal feature consistency loss to optimize both visual and textual prompts concurrently. While incorporating APT methods into FPT can enhance robustness, these approaches do not address a critical challenge in FL: the \emph{class information gap}, which results from data heterogeneity.

Adversarial training follows the ``attack as defense" strategy, where the defender generates adversarial samples to simulate an attack, and then trains the model using these samples to improve its robustness. The effectiveness of this approach depends on the quality of the adversarial samples. In FL, a key challenge is the \textit{class information gap}: each client’s local data only contains a subset of the global dataset. Due to data limitations, clients struggle to generate adversarial samples that can effectively disrupt the full decision boundary. Additionally, the objectives of local clients and the global model are often misaligned. For example, if the global dataset has 100 classes distributed across 10 clients, with each client holding only 10 unique classes, local adversarial training will only optimize decision boundaries for the classes held by that client. As a result, the decision boundaries for the remaining 90 classes will not be optimized. This misalignment further weakens the effectiveness of federated adversarial training. To address this, adversarial training in FL must align the objectives of local clients with those of the global model.

To address this issue, we designed a \textbf{cross-attention-based prompt generator}, which uses the attention mechanism to transform text prompts into visual prompts. Meanwhile, we construct a \textit{beacon} (global label embedding) by aggregating the textual label embeddings from all clients, enabling it to capture global class distributions.  Specifically, the generator uses text prompts, guided by the beacon, to generate the corresponding visual prompts. This allows clients to incorporate data distributions from other clients rather than relying solely on their local data, thus aligning their objectives with the global objective.
Additionally, we propose a \textbf{cross-layer generator sharing} strategy, where a single prompt generator is shared across all layers of the CLIP model. This strategy further enhances the coupling of prompts across layers while reducing the number of trainable parameters. Generator sharing not only improves the model’s ability to generalize across clients but also leads to more efficient training by minimizing redundancy in the model's architecture.

In summary, our main contributions are as follows:
\begin{itemize}
\item We investigate the adversarial robustness of FPT and propose a novel method, \textbf{FedAPT}, to enhance the robustness of FPT-tuned VLMs against potential adversarial attacks, particularly in non-IID settings.

\item We identify a critical challenge in \textbf{FedAPT} caused by data heterogeneity--- \emph{class information gap}---and propose two novel techniques to mitigate this issue: (1) a \textbf{class-aware prompt generator} guided by a global \textit{Beacon}, and (2) \textbf{cross-layer generator sharing}.

\item  We conduct extensive experiments on 15 datasets and demonstrate that \textbf{FedAPT} consistently outperforms five state-of-the-art methods, achieving superior adversarial robustness---notably improving robustness against PGD-100 by 11.49\%. Furthermore, \textbf{FedAPT} maintains higher clean accuracy compared to all competing methods.
\end{itemize}

\vspace{0.2cm}
\section{Related Work}\label{sec:related}
\vspace{0.1cm}
\noindent\textbf{Federated Prompt Tuning.}\; Prompt tuning has been widely applied for fine-tuning VLMs \cite{jia2022visual, khattak2023maple, wang2024lion}. The combination of FL and prompt tuning addresses the challenges of training CLIP's large-scale parameters while offering a novel approach to personalizing large models \cite{zhao2023fedprompt, che2023federated, wei2023dual}. PromptFL \cite{guo2023promptfl} was the first to adapt CoOp \cite{perez2021true} to FL, enabling clients to collaboratively learn unified prompt vectors for efficient global aggregation and local training.  To tackle data heterogeneity, FedPG \cite{yang2023efficient} proposes a personalized FL framework that employs a server-side prompt generator to produce client-specific visual prompts, effectively adapting frozen foundation model backbones to diverse local data distributions. SGPT \cite{deng2024unlocking} addresses the limitations of generalized FL and personalized FL in balancing global and local performance, bridging this gap by combining shared and group-specific prompts. Other works have focused on improving FPT's cross-domain generalization. FedTPG \cite{qiu2024federated} enhances generalization across classes, domains, and datasets by introducing a generator that maps textual class labels to prompts, replacing fixed prompt vectors. Similarly, DiPrompT \cite{bai2024diprompt} separates domain-specific and general features during local training, using distinct lightweight components to minimize interference and eliminate irrelevant information, thereby improving domain generalization. While these studies provide valuable insights into FPT, they overlook a critical issue in FL: robustness against adversarial examples. Addressing this vulnerability is essential for ensuring the security and reliability of federated systems.

\vspace{0.1cm}
\noindent\textbf{Adversarial Prompt Tuning.}\;
Adversarial prompt tuning integrates adversarial training \cite{jia2022adversarial, qian2022survey, kireev2022effectiveness} into visual, textual, or multimodal prompts during prompt tuning, demonstrating its effectiveness in enhancing model robustness. Specifically, AVP \cite{chen2023visual}  improves the adversarial robustness of fixed, pre-trained models during testing by using visual prompts. In contrast, APT \cite{li2024one}and AdvPT \cite{zhang2025adversarial} focus on text prompts to enhance robustness. AdvPT uses fixed adversarial examples to train soft prompts, whereas APT jointly optimizes adversarial examples and adversarial prompts (adv prompts), safeguarding the model against dynamic adversarial attacks. Additionally, AdSPT \cite{wu2022adversarial} extends adversarial prompt tuning techniques to sentiment analysis tasks.
Recent studies suggest that multimodal integration surpasses single-modal approaches, such as vision or text alone. Inspired by Maple \cite{khattak2023maple}, FAP \cite{zhou2024few} introduces a novel training objective that improves multimodal feature consistency while distinguishing unimodal features between natural and adversarial samples. Despite the promising results of these methods, they largely neglect the challenges posed by data heterogeneity in FL.
In this work, we reveal that applying adversarial training in federated heterogeneous environments can cause \textit{class information gap}. To address this issue, we propose FedAPT, the first adversarial prompt tuning method specifically designed for FL.

\section{Proposed Method}\label{sec:fedapt}
\subsection{Preliminaries}
\noindent\textbf{Notations. } We adopt a federated learning framework comprising a central server and $N$ distributed clients. Each client $i$ possesses a private dataset $\mathcal{D}_i = \{(x,y)\}$ with $n_i$ distinct classes, where $x$ denotes images and $y$ their corresponding labels. Following the non-IID setting established in \cite{qiu2024federated}, we assume mutually exclusive class distributions across clients, yielding a total of $C = \sum_{i=1}^N n_i$ unique classes globally. During each communication round, the server uniformly samples $E$ clients to participate in collaborative training. Our architecture builds upon a pre-trained CLIP model, which consists of: (1) an image encoder $\mathcal{I}(\cdot)$ parameterized by $\theta_{\mathcal{I}}$, and (2) a text encoder $\mathcal{T}(\cdot)$ parameterized by $\theta_{\mathcal{T}}$.

\noindent\textbf{CLIP Prompt Tuning.} 
Given an input image $x$, CLIP generates textual embeddings by augmenting class labels with a predefined template (e.g., "a photo of a <class>"). The zero-shot classification probability for class $c$ is computed as:
\begin{equation}
    p(y = c | x) = \frac{\exp \big( \cos(\mathcal{I}(x; \theta_{\mathcal{I}}), \mathcal{T}(w_c; \theta_{\mathcal{T}})) / \tau \big)}{\sum_{i=1}^C \exp \big( \cos(\mathcal{I}(x;\theta_{\mathcal{I}}), \mathcal{T}(w_i; \theta_{\mathcal{T}}))/\tau \big)}, 
\end{equation}
where $\cos(\cdot, \cdot)$ denotes cosine similarity, $C$ represents the total number of classes, $\tau$ is a temperature parameter, and $w_c$ corresponds to the word embedding of the $c$-th class.

In CLIP prompt tuning, we introduce a learnable text prompt $\mathbf{P}$ that modifies the word embeddings. The optimization objective becomes:
\begin{equation}
    \mathbf{P}^* = \argmin_{\mathbf{P}} \mathbb{E}_{(x,y)} \mathcal{L}\big(\cos\big(\mathcal{I}(x;\theta_{\mathcal{I}}), \mathcal{T}([W, \mathbf{P}];\theta_{\mathcal{T}})\big)\big),
\end{equation}
where $\mathcal{L}$ is the cross-entropy loss and $[W, \mathbf{P}]$ denotes the concatenation of base embeddings with the learned prompt.

\noindent\textbf{Adversarial Prompt Tuning.} 
Building upon APT \cite{li2024one}, we generate adversarial examples by adding a bounded perturbation $\boldsymbol{\delta}$ to the input $x$, with $\|\boldsymbol{\delta}\|_p \leq \epsilon$. This perturbation aims to maximize the discrepancy between image and correct text features:
\begin{equation}
    \argmax_{\|\boldsymbol{\delta}\|_p \leq \epsilon} \mathcal{L}\big(\cos\big(\mathcal{I}(x + \boldsymbol{\delta};\theta_{\mathcal{I}}), \mathcal{T}([W, \mathbf{P}];\theta_{\mathcal{T}})\big)\big).
\end{equation}
The robust prompt $\mathbf{P_r}$ is then optimized to maintain performance under such perturbations:
\begin{equation}
\mathbf{P_r}^* = \argmin_{\mathbf{P_r}} \mathbb{E}_{(x,y)} \mathcal{L}\big(\cos\big(\mathcal{I}(x + \boldsymbol{\delta};\theta_{\mathcal{I}}), \mathcal{T}([W,\mathbf{P_r}];\theta_{\mathcal{T}})\big)\big).
\end{equation}

\subsection{FedAPT}
\begin{figure}[ht]
\begin{center}
\centerline{\includegraphics[width=\linewidth ]{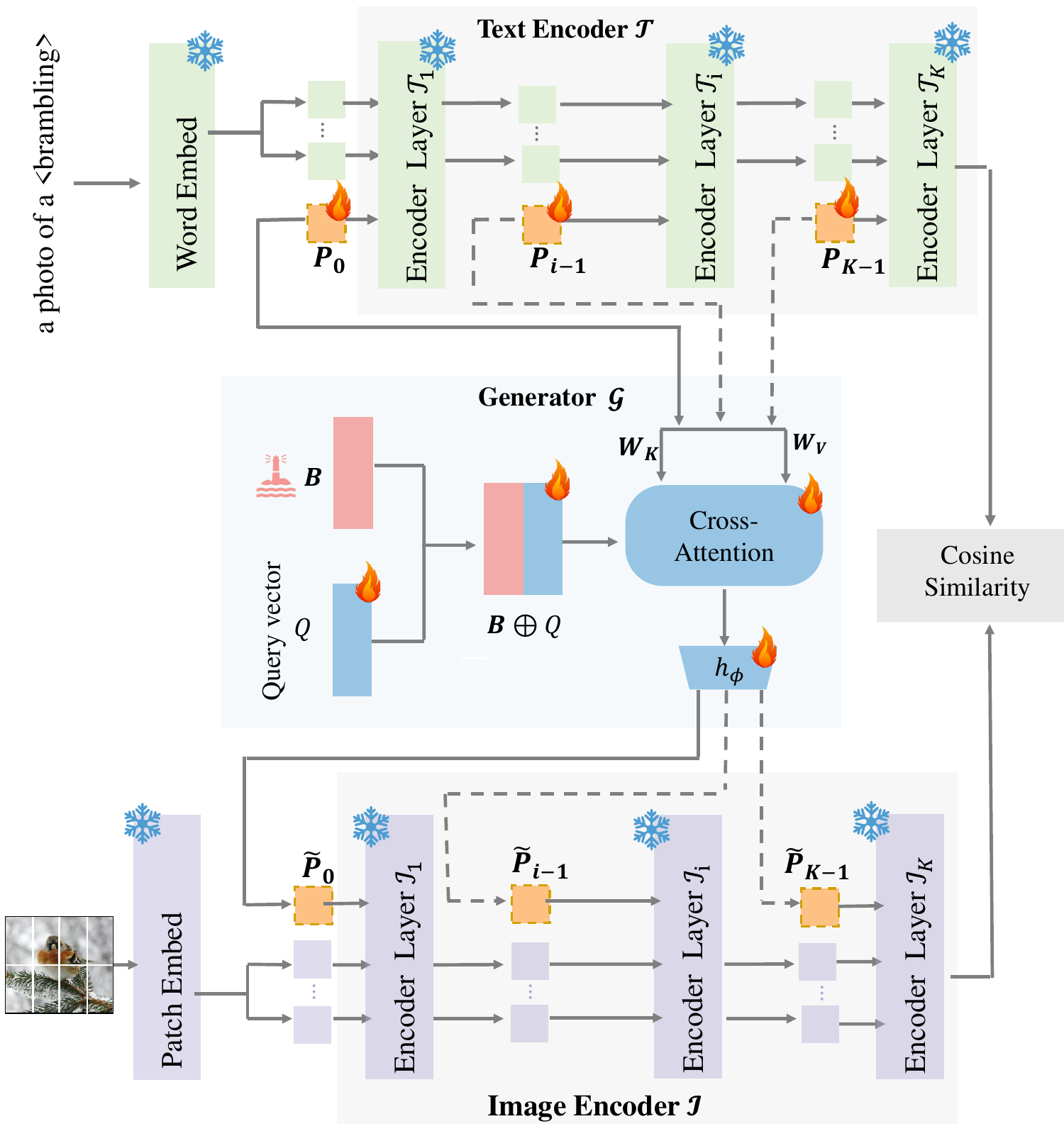}}
\caption{Overview of FedAPT: In this approach, only the text prompt $\mathbf{P}$ and the prompt generator $\bm{\mathcal{G}}$ are tuned, while the rest of the model remains frozen. FedAPT generates vision prompts $\tilde{\mathbf{P}}$ from text prompts $\mathbf{P}$ using the prompt generator $\bm{\mathcal{G}}$, which is guided by $B$. The $B$ token encapsulates the textual label information for all classes. }
\label{Fig: Model}
\end{center}
\vspace{-0.5cm}
\end{figure}

\noindent\textbf{Framework Overview.}\; Figure \ref{Fig: Model} illustrates the architecture of our proposed FedAPT framework. To bridge the label information gap between clients and the global model, caused by data heterogeneity, we leverage the semantic information of both client-specific and global classes. We introduce a class-aware prompt generator that transforms text prompts into visual prompts. Specifically, we design a \textit{beacon} that encodes global class information and integrates it into the prompt generator. This enables the client, during local adversarial training, to focus not only on its own data but also on the global data distribution, aligning local and global objectives. FedAPT consists of two key components: (1) a \textbf{class-aware prompt generator} $\bm{\mathcal{G(\cdot)}}$ which generates visual prompts from text prompts: $\mathbf{\tilde{P}}= \bm{\mathcal{G}}(\mathbf{P})$, guided by a global \textit{Beacon} (denoted as \( B \)), and (2) \textbf{cross-layer generator sharing}. Next, we will introduce the key elements in the two components.

\begin{figure*}[ht]
\begin{center}
\centerline{\includegraphics[width=\linewidth ]{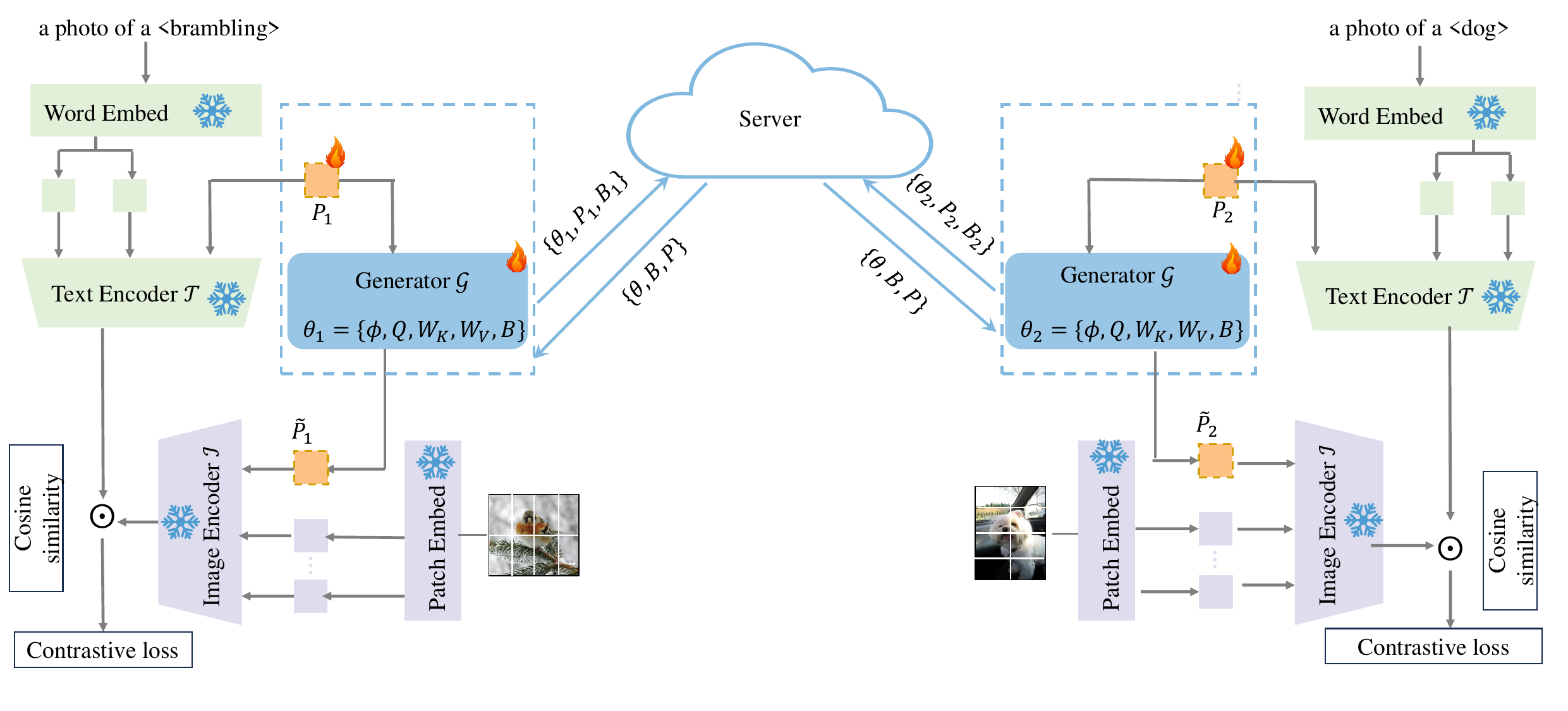}}
\vspace{-0.4cm}
\caption{Our proposed FedAPT collaboratively learns a unified text adversarial prompt $\mathbf{P}$ and a prompt generator $\bm{\mathcal{G}}$ across multiple clients with diverse classification datasets. }
\label{Fig: Framework}
\end{center}
\vspace{-0.5cm}
\end{figure*}

\subsubsection{The Beacon}
In FedAPT, the \emph{beacon} (global label embedding) is a  obtained by aggregating label information across all clients, which serves as a global reference for the class distribution. Rather than sharing raw training data---which would compromise privacy---the beacon generates semantic insights into the global class distribution by performing a weighted aggregation of textual label embeddings (i.e., word embeddings) contributed by each client.

During local training, the \emph{beacon} allows each client to refine its understanding of class semantics by integrating both its local data distribution and the aggregated global view. Specifically, the word embeddings of client \( i \) in communication round \( t \) are represented as \( W_t^i = [ W_t^i, \mathbf{P}] \). The \emph{beacon} is computed by aggregating the last \( m \) token embeddings from each layer of all clients' word embeddings. To ensure stability, the initial beacon \( B_0 \) is derived from clean word embeddings generated using unambiguous prompts (e.g., ``a photo of a <class>”).  

During federated training, the beacon is updated iteratively using a momentum-based approach to balance stability and adaptability:  
\begin{equation}
    B_{t} = \beta B_{t-1} + (1-\beta) \frac{1}{E}\sum_{i=1}^E W_t^i [-m:], 
\end{equation}
where \( \beta \) is a momentum hyper-parameter, \( E \) denotes the number of participating clients, and \( m \) represents the number of prompt tokens appended to each transformer layer.

\subsubsection{Class-aware Prompt Generator}
The class-aware prompt generator leverages a cross-attention mechanism to combine the global \emph{Beacon} with local text prompts, producing visual prompts that maintain alignment with the global class distribution. This fusion enables the generated prompts to incorporate both client-specific features and broader semantic patterns from the federated learning environment.  
The prompt generator \(\bm{\mathcal{G}}\) is implemented as a lightweight cross-attention network with trainable parameters \(\bm{\theta} = \{\bm{\phi}, \mathbf{Q}, \mathbf{W_K}, \mathbf{W_V}\}\). For a given text prompt \(\mathbf{P}\), the visual prompt generation process can be expressed as: 
\begin{equation}
    \bm{\mathcal{G}}(\mathbf{P}) = h_{\bm{\phi}} \left( \text{cross-attention}(B \oplus \mathbf{Q}, \mathbf{W_P}, \mathbf{V_P}) \right),
\end{equation}
where \(\mathbf{W_P}\) and \(\mathbf{V_P}\) are derived from linear projections of the input prompt: \(\mathbf{W_P} = \mathbf{P} \times \mathbf{W_K}\), \(\mathbf{V_P} = \mathbf{P} \times \mathbf{W_V}\); \(\oplus\) represents the concatenation operation between the \emph{Beacon} and the query vector, and  hidden layers $h_{\bm{\phi}}$ project the output of the cross-attention layer to visual prompt $\mathbf{\tilde{P}}$.

This design ensures that the resulting visual prompts are semantically enriched through the integration of global class information from the \emph{Beacon} while preserving relevant local characteristics from the client's text prompts. The cross-attention mechanism facilitates this adaptive fusion, allowing the model to dynamically balance between global and local features during prompt generation.

\subsubsection{Cross-layer Generator Sharing}
FedAPT employs a cross-layer generator sharing strategy to improve efficiency and effectiveness. Rather than maintaining separate prompt generators for each transformer layer --- which would incur significant parameter overhead --- we use a single shared generator across all layers. This unified approach promotes better inter-layer integration while ensuring prompt consistency throughout the network. For the $j$-th transformer block, visual prompts are generated as $\mathbf{\tilde{P}}_j = \bm{\mathcal{G}}(\mathbf{P}_j)$, where $\mathbf{P}_j$ is the layer-specific text prompt and $\bm{\mathcal{G}}$ is the shared generator. The strategy enhances model robustness through improved layer coordination while reducing parameters.

\noindent\textbf{Collaborative Optimization.}\; 
Clients perform local adversarial training while jointly optimizing the text prompt $\mathbf{P}$ and generator $\bm{\mathcal{G}}$ through collaboration. The FedAPT objective is:
\begin{align}
    \min_{(\mathbf{P}, \bm{\theta})} \max_{\| \bm{\delta} \|_p \leq \epsilon}
    \frac{1}{N} \sum_{i=1}^N \mathbb{E}_{(x,y)\in D_i} 
    \mathcal{L}(\cos( z_\mathcal{T}  ,  z_\mathcal{I}  )), 
\end{align}
where $z_\mathcal{T} = \mathcal{T}(W, \mathbf{P}; \theta_\mathcal{T})$ and $z_\mathcal{I} = \mathcal{I}(x+\bm{\delta}, \bm{\mathcal{G}}(\mathbf{P}; \bm{\theta}); \theta_\mathcal{I})$ represent text and image embeddings, respectively.

\subsubsection{Local Training and Server Aggregation}
FedAPT jointly optimizes the text prompt $\mathbf{P}$ and prompt generator $\bm{\mathcal{G}}(\cdot)$ (parameterized by $\bm{\theta} = \{\bm{\phi}, \mathbf{Q}, \mathbf{W_K}, \mathbf{W_V}\}$) through federated adversarial training for multi-client image classification. The training protocol proceeds in three steps per communication round $t$.

\textbf{Step 1.} Each client $i$ downloads the current global text prompt  $\mathbf{P}_t$, prompt generator $\bm{\theta}_t$, and the $B_t$ from the server.

\textbf{Step 2.} For each client $i \in [1,E]$, the local computation starts with adversarial example generation. First, the text prompt $\mathbf{P}_t$ is combined with the client's local word embeddings through concatenation, while the beacon $B_t$ is similarly concatenated with the query vector $Q$. The prompt generator then processes the text prompt to produce a corresponding visual prompt, which integrates with the image patches from the encoder. The adversarial perturbation is generated by solving the following optimization problem:
\begin{equation}
\bm{\delta}^* = \arg\max_{\|\bm{\delta}\|_p\leq\epsilon}\mathcal{L}(\cos(z_\mathcal{T}, z_\mathcal{I})),
\end{equation}
where $z_\mathcal{T} = \mathcal{T}(W,\mathbf{P}_t;\theta_{\mathcal{T}})$ and $z_\mathcal{I} = \mathcal{I}(x+\boldsymbol{\delta}, \mathcal{G}(\mathbf{P}_t;\bm{\theta}_t);\theta_{\mathcal{I}})$ represent the text and image embeddings, respectively.

The adversarial examples are constructed as $\tilde{x} = x + \bm{\delta}^*$. Using these perturbed samples, the client performs adversarial training and updates its local parameters through stochastic gradient descent (SGD) with learning rate $\eta_t$:
\begin{equation}
\mathbf{P}_{t+1}^i, \bm{\theta}_{t+1}^i = \text{SGD}(\eta_t, \bm{\theta}_t, \mathbf{P}_t, \mathcal{L}_i),
\end{equation}
where the loss function $\mathcal{L}_i$ measures the expected cosine similarity between text embeddings $\mathcal{T}(W, \mathbf{P}_t; \theta_\mathcal{T})$ and image embeddings $\mathcal{I}(\tilde{x}, \bm{\mathcal{G}}(\mathbf{P}_t; \bm{\theta}_t); \theta_\mathcal{I})$ over the perturbed examples. Concurrently, the client updates its \emph{Beacon} as $B_{t+1}^i = [W,\mathbf{P}_{t+1}^i][-m:]$.

The updated parameters $\mathbf{P}_{t+1}^i$, $\bm{\theta}_{t+1}^i$, and $B_{t+1}^i$ are then uploaded to the server for aggregation.

\textbf{Step 3.} The server aggregates client updates through federated averaging to compute new global parameters. The text prompt and generator parameters are updated via simple averaging:
\begin{equation}
    \mathbf{P}_{t+1} =\frac{1}{E} \sum_{i=1}^E \mathbf{P}_{t+1}^i , \;\;
    \bm{\theta}_{t+1} = \frac{1}{E} \sum_{i=1}^E  \bm{\theta}_{t+1}^i,
\end{equation}
\begin{equation}\label{eq:beacon}
    B_{t+1} = \beta B_{t} + (1-\beta)\frac{1}{E} \sum_{i=1}^E B_{t+1}^i,
\end{equation}
where $\beta \in (0,1)$ is a hyperparameter controlling the momentum of \textit{Beacon} updates.

\section{Experiments}\label{sec: experiments}

\subsection{Experimental Setup} \label{sec:exp setup}
\noindent\textbf{Datasets.} We employ 15 image datasets including: ImageNet \cite{deng2009imagenet}; \textit{cross-dataset}: Caltech101 \cite{fei2004learning}, OxfordPets \cite{parkhi2012cats}, StanfordCars \cite{krause20133d}, Flowers102 \cite{nilsback2008automated}, Food101 \cite{bossard2014food}, FGVCAircraft \cite{maji2013fine}, SUN397 \cite{xiao2010sun}, UCF101 \cite{soomro2012ucf101}, DTD \cite{cimpoi2014describing}, EuroSAT \cite{helber2019eurosat}; and \textit{cross-domain}:  ImageNetV2 \cite{recht2019imagenet}, ImageNet-Sketch \cite{wang2019learning}, ImageNet-A \cite{hendrycks2021natural}, and ImageNet-R \cite{hendrycks2021many}. We perform adversarial few-shot learning on ImageNet and cross-dataset datasets, and evaluate the zero-shot adversarial robustness of generalizing from ImageNet to downstream datasets (including both cross-dataset and cross-domain datasets). 

\noindent\textbf{Defense Baselines.}\; We compare FedAPT with the following methods: (1) APT, which adapts APT \cite{li2024one} to the FL setting; (2) APT-V, an FL version of VPT \cite{jia2022visual}; (3) APT-VLI, which applies independent visual-language prompting tuning \cite{khattak2023maple} in the FL; (4) APT-VLJ, an FL variant of Maple \cite{khattak2023maple}; and (5) FedTPG \cite{qiu2024federated}, which learns a unified prompt generation network conditioned on task-specific text input. All FL variants of existing prompt tuning methods are implemented using FedAvg \cite{mcmahan2017communication}. All methods are evaluated under the $l_\infty$-norm constraint, using a 3-step PGD with perturbation budget $\epsilon = 1/255$ and step size $\alpha = 2 / 3 \epsilon$, as in \cite{li2024one}.

\noindent\textbf{Robustness Evaluation.}\;  For robustness evaluation, we generate image-based adversarial examples using PGD-100 \cite{madry2017towards}, DI \cite{xie2019improving}, and CW \cite{carlini2017towards}. For PGD-100, we use a perturbation budget of $\epsilon = 1 / 255$, step size $\alpha = 1/4 \epsilon$, and perform 100 attack steps, following the settings in \cite{li2024one}. For DI and CW, we adopt the configurations specified in the TorchAttacks library \cite{kim2020torchattacks}.

\begin{figure*}[!ht]
\begin{center}
\centerline{\includegraphics[width=\linewidth ]{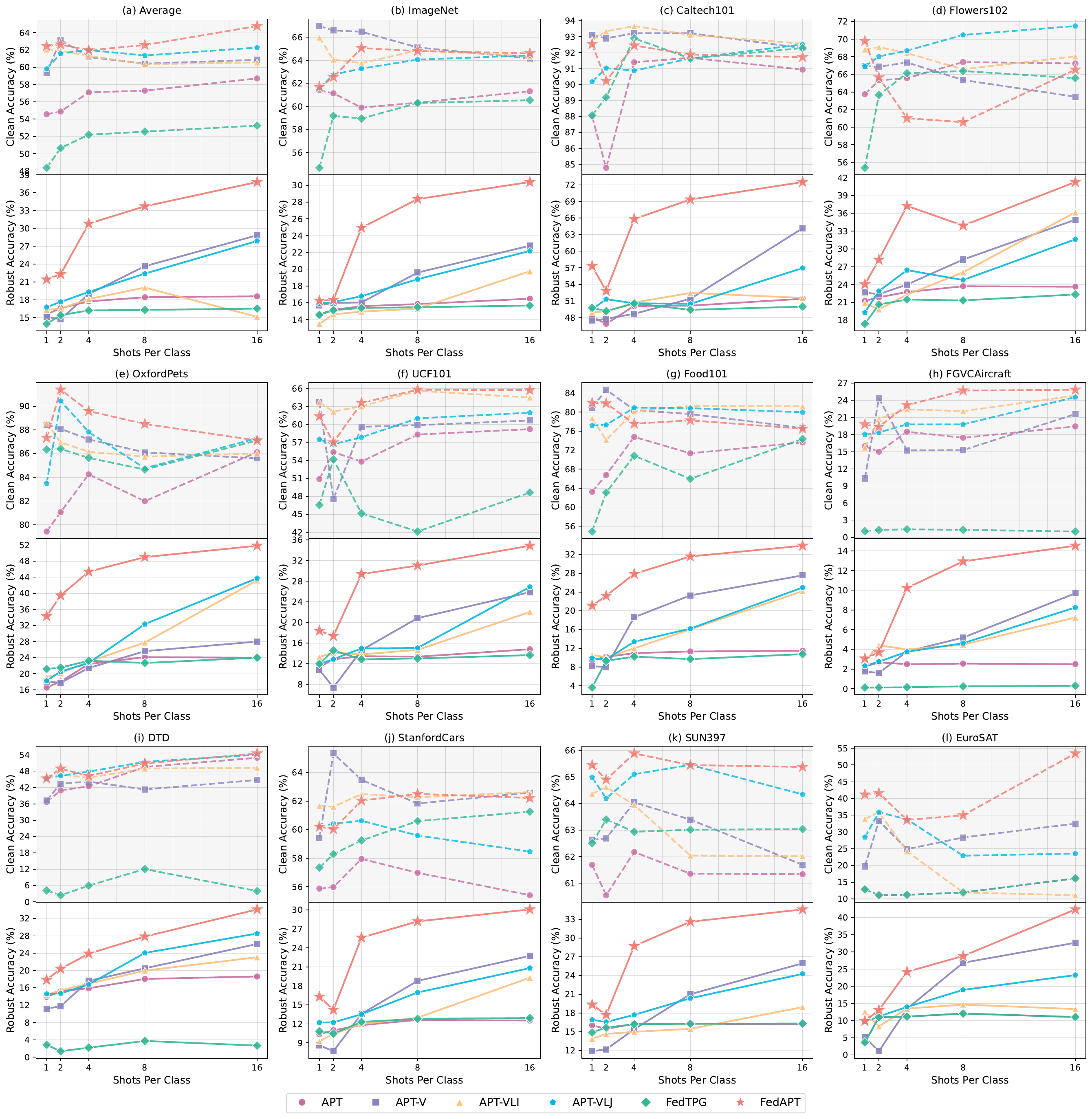}}
\caption{Clean accuracy (\%) and PGD-100 adversarial robustness (\%) across eleven datasets with varying shot numbers [1, 2, 4, 8, 16]. The top figure displays clean accuracy, while the bottom figure shows adversarial robustness. All defense methods (APT, APT-V, APT-VLI, APT-VLJ, FedTPG, and FedAPT) were trained on the eleven datasets using a training perturbation budget of $\epsilon = 1/255$ and $\alpha = 2/3 \epsilon$. 
}
\label{Fig: mian_result}
\end{center}
\vspace{-0.6cm}
\end{figure*}

\begin{table*}[ht]
\caption{Zero-shot adversarial robustness (\%) of various defense methods from ImageNet to downstream datasets (cross-dataset and cross-domain). Evaluations were performed against PGD, DI, and CW under a perturbation budget of $\epsilon = 1/255$. All methods were fine-tuned on ImageNet under an 8-shot setting. The best results are boldfaced, while \underline{\hspace{0.2cm}} indicates the second-best results. Green upward arrows represent improvements in adversarial robustness of FedAPT over the best baseline.}
\vspace{0.1cm}
\label{tab:cross_datasets}
\resizebox{1\linewidth}{!}{
\begin{tabular}{c c c c c c c c c c c c c c c c c | c}
\toprule
 & &\multicolumn{1}{c}{Source} &\multicolumn{14}{c}{Target}\\
        \cmidrule(r){3-3} \cmidrule(r){4-18} 
& & & \multicolumn{11}{c}{Cross - dataset}  & \multicolumn{4}{c}{Cross  - domain} \\
         \cmidrule(r){4-13} \cmidrule(r){14-18} 

Defense & Attack &\rotatebox[origin=c]{60}{\small ImageNet} & \rotatebox[origin=c]{60}{\small Caltech101} & \rotatebox[origin=c]{60}{\small Flowers102} & \rotatebox[origin=c]{60}{\small OxfordPets} & \rotatebox[origin=c]{60}{\small UCF101} & \rotatebox[origin=c]{60}{\small Food101} & \rotatebox[origin=c]{60}{\small FGVCAircraft} & \rotatebox[origin=c]{60}{\small DTD} & \rotatebox[origin=c]{60}{\small StanfordCars} & \rotatebox[origin=c]{60}{\small SUN397}  
         & \rotatebox[origin=c]{60}{\small EuroSAT} & \rotatebox[origin=c]{60}{\small ImageNet-A} & \rotatebox[origin=c]{60}{\small ImageNet-R}  & \rotatebox[origin=c]{60}{\small ImageNet-S}  & \rotatebox[origin=c]{60}{\small ImageNetV2} 
 &\rotatebox[origin=c]{60}{\small Average}\\
\midrule
\multirow{3}{*}{APT} & PGD & 15.85 & 51.11 &  15.75  &  22.56  & 11.12  &	8.67  &  1.83  &   12.11   &  9.67  &  16.68 &	\underline{8.14}  &       3.49 & 36.46 & 26.88 & 12.81  & 16.87  \\
& DI & 20.49 & 61.66 &  19.01  &  29.87  &	  16.17  &	13.56  &  3.54  &   15.66   &  13.23  &  22.13 & \underline{8.82}     & 4.69 & 41.64 & 28.52 & 16.52    &  21.03  \\
& CW & 2.73 & 16.18 &  3.29  &  3.13  &	  1.03  &	0.52  &  1.35  &   5.79   &  1.74  &  3.07 &	\underline{7.86}    & 0.31 & 13.29 & 12.29 & 2.05  &  4.97  \\
\midrule

\multirow{3}{*}{APT-V} & PGD & \underline{19.60} & \underline{61.98} &  \underline{25.09}  &  
\underline{31.39}  &	\underline{17.10}  & \underline{13.14}  & \underline{5.01}  &   \underline{16.07}   &  \underline{14.70}  &  \underline{21.96} &	4.07  &  3.68 & \underline{40.16} & \underline{28.92} & \underline{15.64}    &  \underline{21.24}  \\
& DI & \underline{23.29} & \underline{66.28} &  28.05  &  \underline{37.06}  &	 \underline{20.85}  &	\underline{17.02}  &  \underline{7.62}  &   \underline{17.96}   &  \underline{18.24}  &  \underline{25.91} &	4.95   & 4.74 & \underline{43.57} & \underline{29.93} & 18.64  &  \underline{24.28}  \\
& CW & \underline{4.44} & \underline{24.82} &  \underline{9.17}  &  \underline{4.96}  &	  \underline{1.51}  &	\underline{1.35}  &  \underline{1.35}  &   \underline{8.45}   &  \underline{3.05}  &  \underline{6.09} &	\underline{1.23}  & \underline{0.4} & \underline{16.95} & \underline{17.19} & \underline{3.10}  &  \underline{6.93}  \\
\midrule

\multirow{3}{*}{APT-VLI} & PGD &  15.31 & 52.57 &  22.12  &  21.01  &	11.04  & 9.47  &  1.77  &   13.35   &  10.43  &  16.44 & 4.60  & 3.22 & 36.51 & 27.74 & 11.97 &  17.17 \\
& DI & 20.35 & 64.86 &  26.39  &  28.50  &	  18.37  &	14.19  &  4.8  &   17.13   &  14.92  &  22.49 & 5.87 & 4.57 & 41.71 & 29.81 & 17.25  &  22.08  \\
& CW & 1.74 & 11.88 &  3.49  &  1.77  &	  0.21  &	0.26  &  0.84  &   3.84   &  1.16  &  2.43 &	0.84 & 0.16 & 10.51 & 11.30 & 1.39 &  3.45  \\
\midrule

\multirow{3}{*}{APT-VLJ} & PGD & 18.81 & 54.32 &  23.30  &  27.20  & 12.45  &	10.46  &  2.28  &   13.65   &  11.18  &  18.50 &	3.49  & \underline{3.78} & 37.78 & 27.94 & 14.31 &  18.59  \\
& DI & 23.26 & 62.19 &  \underline{28.90}  &  35.37  &	17.26  &  15.41 &  4.32  &   16.66   &  14.82  &  24.57 &	4.17 & \underline{5.57} & 42.71 & 29.62 & \underline{18.84} &  22.91  \\
& CW & 3.40 & 16.91 &  4.87  &  \underline{5.26}  &	  0.92  &	0.70  &  0.84  &   6.02   &  1.95  &  3.65 &	0.07 & 0.29 & 14.01 & 14.00 & 2.51  &  5.02  \\
\midrule

\multirow{3}{*}{FedTPG} &PGD & 15.47 & 50.02 &  11.40  &  19.18  &	 11.60  &	8.46  &  1.65  &   13.35   &  7.61  &  16.33 &	0.46 & 3.30 & 36.3 & 27.02 & 12.21 &  15.62 \\
& DI & 20.41 & 59.43 &  13.47  &  29.89  &	  17.65  &	13.64  &  3.18  &   15.83   &  11.50  &  21.34 &	0.64 & 4.97 & 41.57 & 28.79 & 16.99   &  19.95  \\
& CW & 2.71 & 16.26 &  2.76  &  3.46  &	  0.79  &	0.77 &  0.48  &   5.96   &  0.82  &  2.97 &	0.11  & 0.34 & 13.05 & 12.01 & 1.9 & 4.29\\
\midrule

\multirow{6}{*}{\textbf{FedAPT}} & PGD & \textbf{28.36} & \textbf{71.36} &  \textbf{34.72}  &  \textbf{48.87}  & \textbf{24.13}  & \textbf{20.13}   &  \textbf{6.57}   &   \textbf{20.86}   &  \textbf{21.01}  &  \textbf{29.52}  &	\textbf{8.24}  & \textbf{6.13}  & \textbf{46.05}  & \textbf{28.94}   & \textbf{22.78}  &  \textbf{27.84}  \\

& &  \textcolor{green}{ ( 8.96 $\uparrow$)} &  \textcolor{green}{ (9.38 $\uparrow$)}  &  \textcolor{green}{ (9.62 $\uparrow$)}  &  \textcolor{green}{ (17.47 $\uparrow$)}  & \textcolor{green}{ (7.03 $\uparrow$)}  &  \textcolor{green}{ (6.99 $\uparrow$)}  & \textcolor{green}{ (1.56 $\uparrow$)}  & \textcolor{green}{ (4.79 $\uparrow$)}  &  \textcolor{green}{ (6.31 $\uparrow$)} &  \textcolor{green}{ (7.56 $\uparrow$)}  & \textcolor{green}{ (0.1 $\uparrow$)}   &   \textcolor{green}{ (2.35$\uparrow$)}  &  \textcolor{green}{ (5.89$\uparrow$)}   &   \textcolor{green}{ (0.02$\uparrow$)}  &  \textcolor{green}{ (7.14$\uparrow$)} & \textcolor{green}{ (6.60 $\uparrow$)}  \\

& DI & \textbf{31.64}  & \textbf{73.79}   &  \textbf{36.87}   & \textbf{52.58}   &	 \textbf{27.25}  &	\textbf{23.39}  &  \textbf{8.13}   &   \textbf{23.17}  &  \textbf{24.61}  &  \textbf{32.93}   &	\textbf{9.69}  & \textbf{7.52}   & \textbf{48.81} &  \textbf{29.97}  & \textbf{24.98}   &  \textbf{30.32}   \\
& & \textcolor{green}{ (8.35 $\uparrow$)} &   \textcolor{green}{ (7.51$\uparrow$)}  &   \textcolor{green}{ (7.97 $\uparrow$)}  &   \textcolor{green}{ (15.52 $\uparrow$)}   &   \textcolor{green}{ (6.40 $\uparrow$)}  &  \textcolor{green}{ (6.37 $\uparrow$)}  &  \textcolor{green}{ (0.51 $\uparrow$)}  & \textcolor{green}{ (5.21 $\uparrow$)}  &  \textcolor{green}{ (6.37 $\uparrow$)}  & \textcolor{green}{ (7.02 $\uparrow$)}  & \textcolor{green}{ (0.87 $\uparrow$)} &   \textcolor{green}{ (1.95$\uparrow$)}  &  \textcolor{green}{ (5.24$\uparrow$)}   &   \textcolor{green}{ (0.04$\uparrow$)}  &  \textcolor{green}{ (6.14$\uparrow$)} &  \textcolor{green}{ (6.03 $\uparrow$)} \\

& CW & \textbf{15.63} & \textbf{45.08} &  \textbf{22.49}  &  \textbf{27.58}  &	  \textbf{9.27}  &	\textbf{8.43} &  \textbf{3.87}  &   \textbf{16.84}   &  \textbf{12.13}  &  \textbf{18.17} &	\textbf{7.95}  & \textbf{2.04} & \textbf{30.09} & \textbf{21.77} & \textbf{11.91} &  \textbf{16.92} \\
& & \textcolor{green}{ (11.19 $\uparrow$)} &   \textcolor{green}{ (20.25 $\uparrow$)}  &   \textcolor{green}{ (13.32 $\uparrow$)}  &   \textcolor{green}{ (22.61 $\uparrow$)}   &   \textcolor{green}{ (7.76 $\uparrow$)}  &  \textcolor{green}{ (7.08 $\uparrow$)}  &  \textcolor{green}{ (2.52$\uparrow$)}  & \textcolor{green}{ (8.39 $\uparrow$)}  &  \textcolor{green}{ (9.08 $\uparrow$)}  & \textcolor{green}{ (12.08$\uparrow$)}  & \textcolor{green}{ (0.09$\uparrow$)} &   \textcolor{green}{ (1.64$\uparrow$)}  &  \textcolor{green}{ (13.14$\uparrow$)}   &   \textcolor{green}{ (4.58$\uparrow$)}  &  \textcolor{green}{ (8.81$\uparrow$)} &   \textcolor{green}{ (9.98 $\uparrow$)} \\
\bottomrule
\end{tabular}}
\end{table*}

\noindent\textbf{Implementation Details.}\;
All methods use a frozen CLIP model with ViT-B/16 as the backbone. Each transformer block contains 2 prompt tokens ($m = 2$) and a depth of 8 ($J = 8$), ensuring a consistent total number of prompt tokens across all methods. For FedAPT, we design a prompt generator with a four-head cross-attention layer (with layer norm) and a linear layer ($h_\theta$) with ReLU. The dimensions of $Q$, $W_K$, and $W_V$ are all set to 512, while the linear layer $h_\theta$ has dimensions of $512 \times 768$. The $B$ (\textit{beacon}) has a dimension of 512.

For FL setting, each client is assigned 10 unique, non-overlapping data classes (with 8 samples per class), resulting in a non-IID distribution, as described in FedTPG \cite{qiu2024federated}. For cross-dataset and cross-domain generalization evaluations, the model is trained on ImageNet and tested on external datasets using zero-shot testing. The training lasts for over 30 rounds, with 1 local epoch per round, using the SGD optimizer (momentum: 0.9) and an initial learning rate of 0.0035. A cosine learning rate scheduler with warm-up is applied during the first epoch, and the momentum $\beta$ in Eq. \eqref{eq:beacon} is set to 0.9. The participation rate is set to $E / N = 1$.

\subsection{Main Results} \label{sec:main result}

\textbf{Adversarial Few-shot Learning. }\; 
We evaluate the adversarial robustness and clean accuracy of defense methods in few-shot adversarial training across eleven datasets, with per-class sample sizes of [1, 2, 4, 8, 16]. As shown in Figure \ref{Fig: mian_result}, FedAPT consistently outperforms five baseline methods in both PGD-100 adversarial robustness and clean accuracy. The results reveal three key trends: First, FedAPT achieves substantial improvements in robustness, with average gains of \textbf{4.62\%}, \textbf{4.67\%}, \textbf{11.49\%}, \textbf{11.33\%}, and \textbf{8.96\%} for the [1, 2, 4, 8, 16]-shot settings, respectively. These improvements scale positively with the sample size. Second, modality comparisons highlight distinct patterns: while text prompt methods (APT, FedTPG) underperform in federated heterogeneous environments and show limited adaptability to sample size variations, visual prompt methods (APT-V) demonstrate stronger performance and are better suited for FL. Dual-modal (text-visual) approaches (APT-VLI, APT-VLJ) offer only marginal gains in robustness, suggesting that simply combining text and vision modalities does not address the unique challenges of FL — a gap that FedAPT’s integrated design successfully overcomes. Finally, FedAPT not only improves adversarial robustness but also maintains superior clean accuracy, achieving an optimal balance between defense and task performance. These results demonstrate the effectiveness of FedAPT in data-scarce scenarios, where its text-to-visual joint prompt tuning strategy addresses the constraints of FL, providing state-of-the-art robustness without compromising clean accuracy.

\begin{table*}[!ht]
\caption{Zero-shot adversarial robustness (\%) transferring from ImageNet to downstream datasets (cross-dataset and cross-domain) under PGD-100, DI and CW attacks, comparing generator sharing (Share) with independent generators (Inde). Each client has 20 non-overlapping classes. }
\label{tab:generator_share}
\resizebox{0.97\linewidth}{!}{
\begin{tabular}{ccccccccccccccccc|c}
\toprule
& &\multicolumn{1}{c}{Source} &\multicolumn{14}{c}{Target}\\
\cmidrule(r){3-3} \cmidrule(r){4-18} 
& & & \multicolumn{11}{c}{Cross - dataset}  & \multicolumn{4}{c}{Cross  - domain} \\
\cmidrule(r){4-13} \cmidrule(r){14-18}
Attack & &\rotatebox[origin=c]{45}{\small ImageNet} & \rotatebox[origin=c]{45}{\small Caltech101} & \rotatebox[origin=c]{45}{\small Flowers102} & \rotatebox[origin=c]{45}{\small OxfordPets} & \rotatebox[origin=c]{45}{\small UCF101} & \rotatebox[origin=c]{45}{\small Food101} & \rotatebox[origin=c]{45}{\small FGVCAircraft} & \rotatebox[origin=c]{45}{\small DTD} & \rotatebox[origin=c]{45}{\small StanfordCars} & \rotatebox[origin=c]{45}{\small SUN397}  
& \rotatebox[origin=c]{45}{\small EuroSAT} 
& \rotatebox[origin=c]{45}{\small ImageNet-A}
& \rotatebox[origin=c]{45}{\small ImageNet-R}
& \rotatebox[origin=c]{45}{\small ImageNet-S} 
& \rotatebox[origin=c]{45}{\small ImageNetV2} 
&\rotatebox[origin=c]{45}{\small Average}\\
\cmidrule(r){1-3} \cmidrule(r){4-13} \cmidrule(r){14-18}
\multirow{3}{*}{\textbf{PGD}} & Inde &  30.19 &  71.92  &  37.15 &  50.77  &	24.71  & 22.30 & 6.48  &  21.15  & 22.74  & 30.37 &  8.13 &  6.36  & 46.77 & 28.99 & 24.26 & 28.82 \\
& Share &  \textbf{31.03} &  \textbf{73.14}   &  \textbf{37.92}  &  \textbf{51.79}  &	\textbf{25.64}  & \textbf{22.79}  & \textbf{6.81}  &  \textbf{22.34}  & \textbf{23.04}  & \textbf{31.33} &  \textbf{8.49} &  \textbf{6.41} & \textbf{47.07} & \textbf{29.12} &  \textbf{24.86} & \textbf{29.44}  \\
& & \textcolor{green}{ (0.83$\uparrow$)}
& \textcolor{green}{ (1.22 $\uparrow$)}
& \textcolor{green}{ (0.77 $\uparrow$)}
& \textcolor{green}{ (1.02 $\uparrow$)}
& \textcolor{green}{ (0.93 $\uparrow$)}
& \textcolor{green}{ (0.49 $\uparrow$)}
& \textcolor{green}{ (0.33 $\uparrow$)}
& \textcolor{green}{ (1.19 $\uparrow$)}
& \textcolor{green}{ (0.30 $\uparrow$)}
& \textcolor{green}{ (0.96 $\uparrow$)}
& \textcolor{green}{ (0.36 $\uparrow$)}
& \textcolor{green}{ (0.05 $\uparrow$)}
& \textcolor{green}{ (0.29 $\uparrow$)}
& \textcolor{green}{ (0.13 $\uparrow$)}
& \textcolor{green}{ (0.40 $\uparrow$)}
& \textcolor{green}{ (0.62 $\uparrow$)}\\
\midrule
\multirow{3}{*}{\textbf{DI}} & Inde &  33.18 &  75.01  &  38.77 &  54.37  &	28.09  & 25.37 & 7.89  &  22.87  & 25.86  & 33.53 &  7.69 &  7.29 & 49.20 & 29.99 & 26.62 & 31.04 \\
& Share &  \textbf{33.87} &  \textbf{75.62}   &  \textbf{39.46}  &  \textbf{55.22}  &	\textbf{28.42}  & \textbf{25.73}  & \textbf{8.16}  &  \textbf{23.64}  & \textbf{26.25}  & \textbf{34.18} &  \textbf{8.79} &  \textbf{7.64} & \textbf{49.48} & \textbf{30.04} &  \textbf{27.10} & \textbf{31.57}  \\
& & \textcolor{green}{ (0.69$\uparrow$)}
& \textcolor{green}{ (0.51 $\uparrow$)}
& \textcolor{green}{ (0.69 $\uparrow$)}
& \textcolor{green}{ (0.85 $\uparrow$)}
& \textcolor{green}{ (0.33 $\uparrow$)}
& \textcolor{green}{ (0.36 $\uparrow$)}
& \textcolor{green}{ (0.27 $\uparrow$)}
& \textcolor{green}{ (0.77 $\uparrow$)}
& \textcolor{green}{ (0.39 $\uparrow$)}
& \textcolor{green}{ (0.65 $\uparrow$)}
& \textcolor{green}{ (1.10 $\uparrow$)}
& \textcolor{green}{ (0.35 $\uparrow$)}
& \textcolor{green}{ (0.28 $\uparrow$)}
& \textcolor{green}{ (0.05 $\uparrow$)}
& \textcolor{green}{ (0.48 $\uparrow$)}
& \textcolor{green}{ (0.53 $\uparrow$)}\\
\midrule
\multirow{3}{*}{\textbf{CW}} & Inde &  17.38 &  49.45  &  25.05 &  31.45  &	11.57  & 10.00 & 3.33  &  17.08  & 13.81  & 19.12 &  7.80 &  2.41 & 31.06 & 22.45 & 13.67 & 18.37 \\
& Share &  \textbf{18.82} &  \textbf{51.85}   &  \textbf{27.69}  &  \textbf{33.14}  &	\textbf{12.93}  & \textbf{10.61}  & \textbf{3.78}  &  \textbf{18.44}  & \textbf{13.95}  & \textbf{20.60} &  \textbf{8.95} &  \textbf{2.45} & \textbf{32.23} & \textbf{22.74} &  \textbf{14.96} & \textbf{19.01}  \\
& & \textcolor{green}{ (1.44$\uparrow$)}
& \textcolor{green}{ (2.40 $\uparrow$)}
& \textcolor{green}{ (2.64 $\uparrow$)}
& \textcolor{green}{ (1.69 $\uparrow$)}
& \textcolor{green}{ (1.36 $\uparrow$)}
& \textcolor{green}{ (0.61 $\uparrow$)}
& \textcolor{green}{ (0.45 $\uparrow$)}
& \textcolor{green}{ (1.36 $\uparrow$)}
& \textcolor{green}{ (0.14 $\uparrow$)}
& \textcolor{green}{ (1.48 $\uparrow$)}
& \textcolor{green}{ (0.15 $\uparrow$)}
& \textcolor{green}{ (0.04 $\uparrow$)}
& \textcolor{green}{ (1.17 $\uparrow$)}
& \textcolor{green}{ (0.29 $\uparrow$)}
& \textcolor{green}{ (1.29 $\uparrow$)}
& \textcolor{green}{ (1.11 $\uparrow$)}\\
\bottomrule
\end{tabular}
}
\end{table*}

\begin{figure*}[!ht]
\begin{center}
    \begin{minipage}[b]{0.33\linewidth}
        \includegraphics[width=\linewidth]{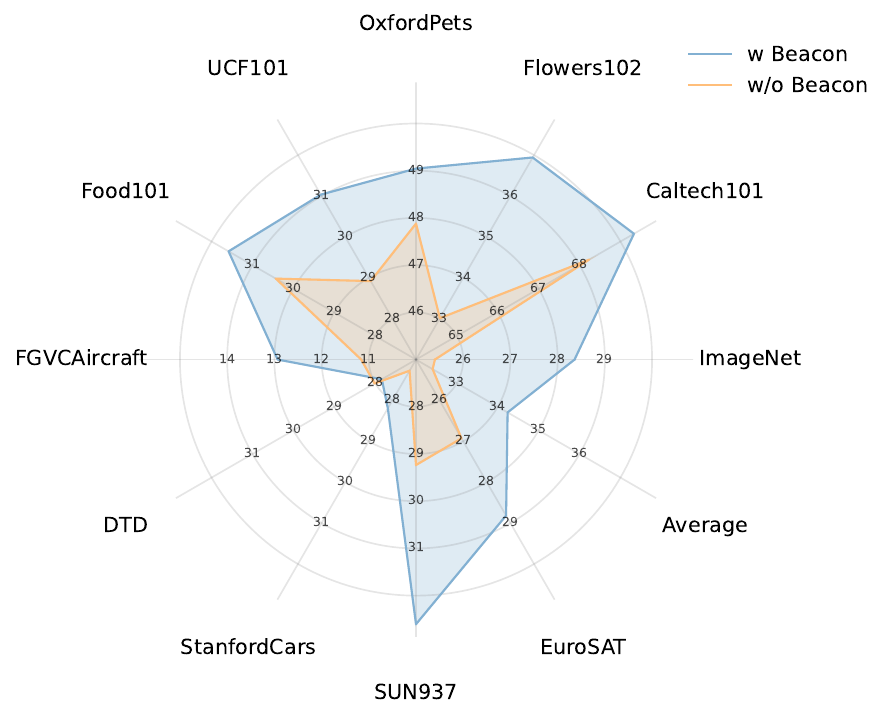}
        \centerline{(a)}
    \end{minipage}
    \hfill
    \begin{minipage}[b]{0.33\linewidth}
        \includegraphics[width=\linewidth]{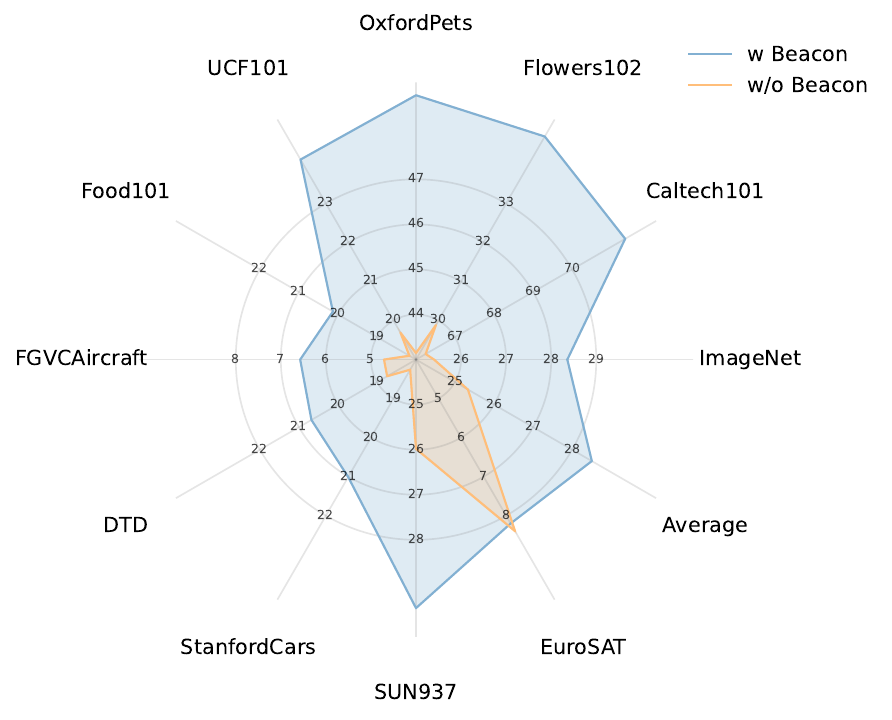}
        \centerline{(b)}
    \end{minipage}
    \hfill
    \begin{minipage}[b]{0.33\linewidth}
        \includegraphics[width=\linewidth]{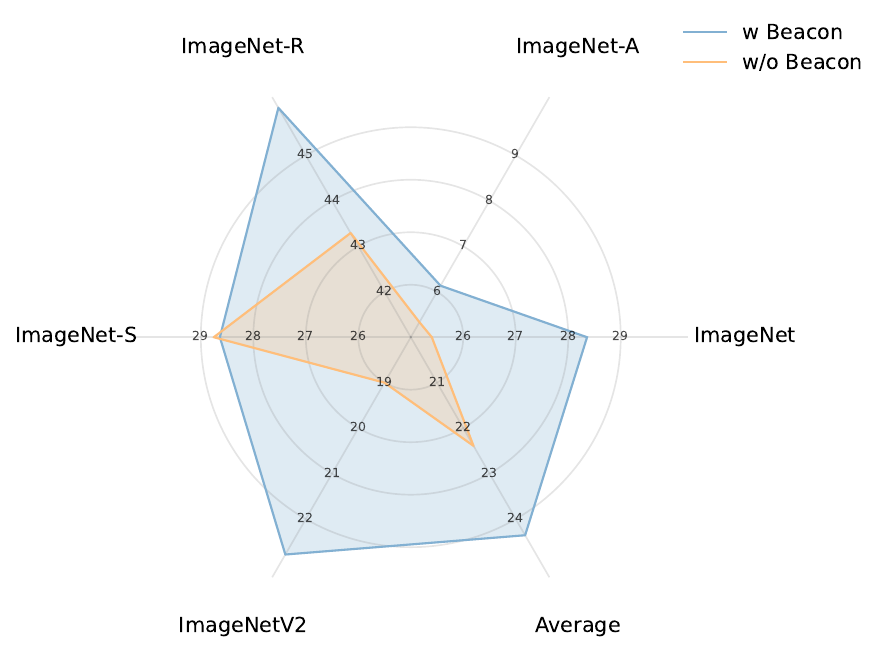}
        \centerline{(c)}
    \end{minipage}
    \vspace{-0.3cm}
    \caption{Comparison of PGD-100 adversarial robustness (\%) for FedAPT with and without \textit{Beacon} guidance. (a) Results on training and testing within the same dataset across 11 datasets; (b) Zero-shot adversarial robustness (\%) from ImageNet to cross-dataset datasets; (c) Zero-shot adversarial robustness (\%) transferring from ImageNet to cross-domain datasets. `w Beacon' refers to models with guidance, while `w/o Beacon' refers to models without guidance.}
    \label{Fig: Impact_beacon}
\end{center}
\vspace{-0.3cm}
\end{figure*}

\noindent\textbf{Generalization to Unseen Datasets. }\; 
We evaluate the model's generalization capability through zero-shot adversarial testing, transferring from ImageNet to unseen datasets in both cross-dataset and cross-domain scenarios. Table \ref{tab:cross_datasets} shows that FedAPT outperforms baseline methods in adversarial robustness against three attack methods (PGD-100, DI, and CW) on 14 downstream datasets (10 cross-dataset and 4 cross-domain datasets). FedAPT achieves average robustness improvements of \textbf{6.60\%}, \textbf{6.03\%}, and \textbf{9.98\%} over the strongest baselines under PGD-100, DI, and CW attacks, respectively. Notably, FedAPT achieves significant gains on OxfordPets, with improvements of \textbf{17.47\%}, \textbf{15.52\%}, and \textbf{22.61\%} against each attack. FedAPT also demonstrates consistent advantages in cross-domain settings, outperforming all baselines, including APT-V, which exhibits the best performance among the baseline methods. Overall, these results highlight FedAPT's enhanced ability to generalize across diverse data distributions while maintaining robustness against a variety of adversarial threats.

\subsection{Ablation Studies} \label{sec:Ablation}
\noindent\textbf{With vs. Without \textit{Beacon}. }\;
To evaluate the importance of the beacon, we compared model performance with and without beacon-guided prompt generation. Figure \ref{Fig: Impact_beacon} highlights the following three key findings. (1) \emph{Enhanced adversarial robustness}: Figure \ref{Fig: Impact_beacon}(a) shows a \textbf{1.83\%} average improvement in PGD-100 robustness across 11 datasets. Notable gains include \textbf{3.91\%} on Flowers102 and \textbf{3.37\%} on SUN397, demonstrating the beacon’s ability to enhance model resilience against adversarial attacks. (2) \emph{Cross-dataset generalization}: Figure \ref{Fig: Impact_beacon}(b) reveals a \textbf{3.17\%} average increase in robustness for zero-shot transfer from ImageNet to cross-dataset tasks. Significant improvements are seen on Caltech101 (\textbf{+5.11\%}), Flowers102 (\textbf{+4.83\%}), OxfordPets (\textbf{+5.73\%}), and UCF101 (\textbf{+4.44\%}), emphasizing the beacon’s role in generalizing learned features across datasets. (3) \emph{Cross-domain adaptation}: Figure \ref{Fig: Impact_beacon}(c) demonstrates consistent robustness gains when transferring from ImageNet to cross-domain datasets, confirming the beacon’s effectiveness in bridging domain gaps. These results validate that beacon-guided global class information in federated heterogeneous settings strengthens both adversarial robustness and generalization across datasets and domains. The beacon guidance mechanism aligns local model updates with global feature distributions, mitigating performance degradation caused by heterogeneity.

\begin{figure*}[ht]
\begin{center}
\centerline{\includegraphics[width=\linewidth ]{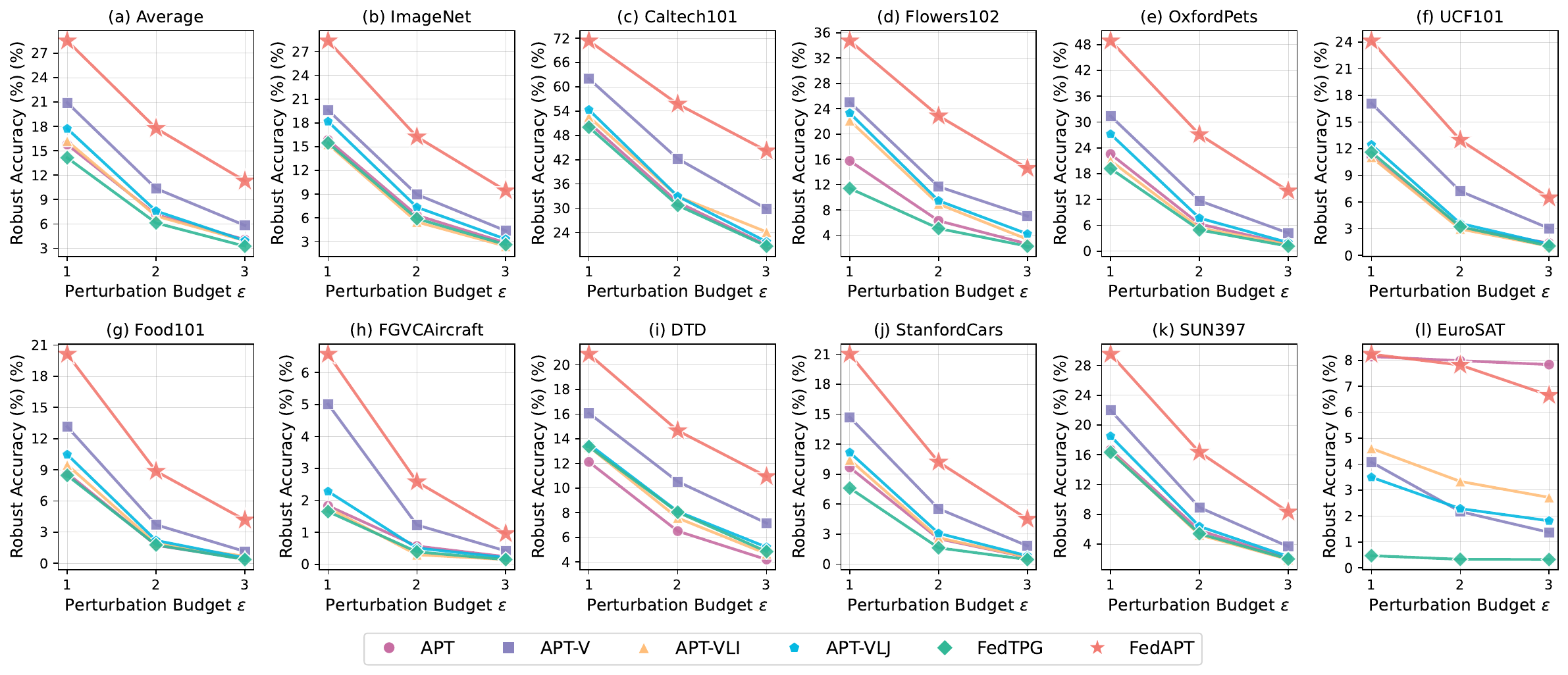}}
\caption{Zero-shot PGD-100 adversarial robustness (\%) from ImageNet to corss-dataset datasets , under perturbation budgets of $\epsilon = 1/255$, $2/255$, and $3/255$. }
\label{Fig: eps}
\end{center}
\vspace{-0.4cm}
\end{figure*}

\begin{figure}[ht]
\begin{center}
\centerline{\includegraphics[width=\linewidth ]{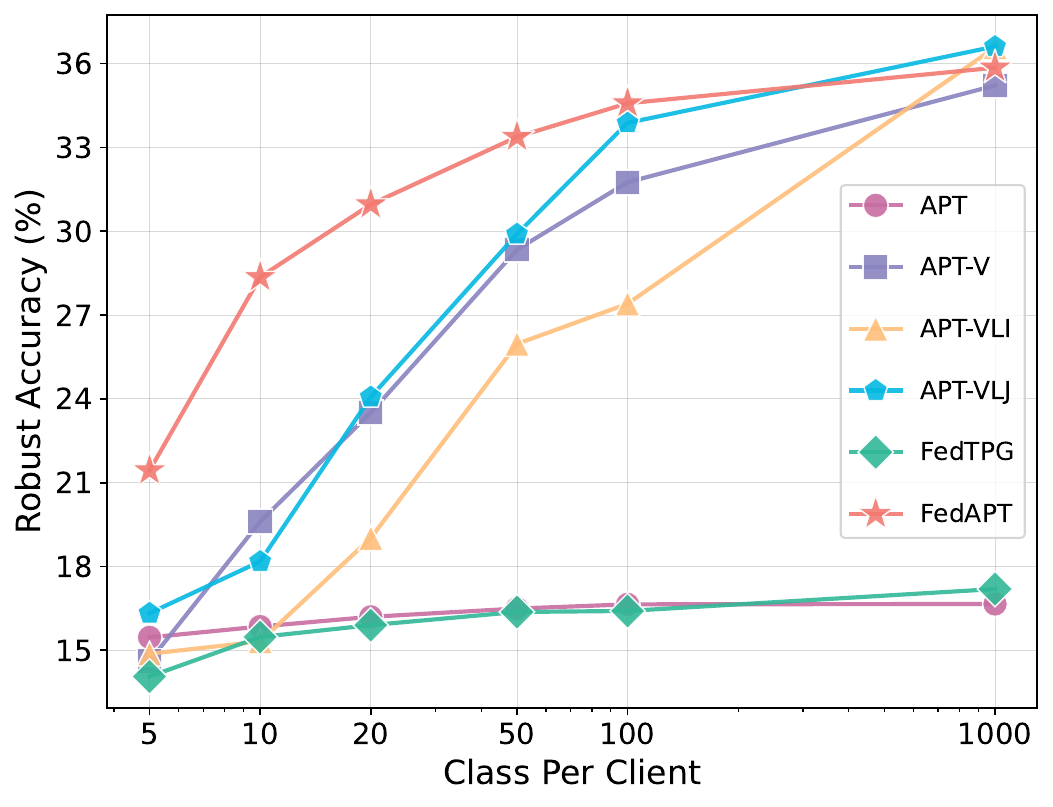}}
\caption{PGD-100 adversarial robustness (\%) with varying numbers of classes per client ([5, 10, 20, 50, 100, 1000]) on ImageNet. }
\label{Fig: noniid}
\end{center}
\vspace{-0.5cm}
\end{figure}

\noindent\textbf{Generator Sharing vs. Independent Generators. }\; 
Table \ref{tab:generator_share} compares the robustness of cross-layer shared generators and independent layer-wise generators under PGD-100, DI, and CW adversarial attacks. The evaluation includes ImageNet as well as cross-dataset and cross-domain downstream tasks, with each client assigned 20 mutually exclusive categories to control computational costs. The results show that the shared generator significantly boosts model robustness by strengthening inter-layer coupling. On average, the generator sharing improves robustness by \textbf{0.62\%} (PGD-100), \textbf{0.53\%} (DI), and \textbf{1.11\%} (CW) across the three attacks. Notable performance gains under specific attacks include a \textbf{1.22\%} improvement for PGD-100 on Caltech101, a \textbf{1.10\%} improvement for DI on EuroSAT, and for CW, \textbf{1.44\%} on ImageNet, \textbf{2.40\%} on Caltech101, \textbf{2.64\%} on Flowers102, and \textbf{1.69\%} on OxfordPets.
These findings indicate that cross-layer generator sharing not only reduces the model's parameter size but also enhances adversarial robustness by fostering stronger inter-layer coupling. This strategy effectively balances the need for model efficiency with improved attack resistance.

\noindent\textbf{Effect of Data Heterogeneity. }\;
Data heterogeneity presents a critical challenge in FL. To assess its impact, we simulated varying levels of heterogeneity by assigning clients different numbers of classes in ImageNet: fewer classes per client (e.g., [5, 10, 20, 50, 100]) represent high heterogeneity, while full-class access (1000 classes) mimics a centralized learning scenario. As shown in Figure \ref{Fig: noniid}, two key patterns emerge. First, as the number of classes per client increases (reducing data heterogeneity), PGD-100 adversarial robustness improves significantly. FedAPT outperforms baseline methods in high-heterogeneity scenarios, with robustness gains of \textbf{5.11\%}, \textbf{8.75\%}, \textbf{6.89\%}, \textbf{3.48\%}, and \textbf{0.7\%} for the [5, 10, 20, 50, 100] class settings, respectively. Second, in the centralized setting with full-class access (1000 classes), FedAPT performs similarly to or slightly worse than methods like APT-VLI and APT-V. This is because centralized learning eliminates class information gap by providing direct access to the global data distribution, reducing FedAPT's advantage in addressing heterogeneity-induced disparities. These results highlight FedAPT’s effectiveness in federated heterogeneous environments and underscore its alignment with the core FL principle: leveraging decentralized, partial data views to build robust models.

\noindent\textbf{Effect of Perturbation Budget $\epsilon$. }\;
This experiment investigates how different perturbation budgets ($\epsilon = 1/255, 2/255, 3/255$) impact model robustness through zero-shot PGD-100 adversarial testing, transferring from ImageNet to cross-dataset tasks. As shown in Figure \ref{Fig: eps}, model robustness decreases progressively with larger $\epsilon$ values, a trend observed across all 11 datasets. Despite this, FedAPT consistently outperforms baseline methods, with average robustness improvements of \textbf{7.61\%} ($\epsilon = 1/255$), \textbf{7.41\%} ($\epsilon = 2/255$), and \textbf{5.47\%} ($\epsilon = 3/255$) over the strongest baseline. Notably, while higher adversarial strength ($\epsilon = 3/255$) leads to reduced absolute performance for all models, FedAPT maintains its advantage, demonstrating strong defense capabilities even under more intense attacks. These results underscore FedAPT's effectiveness in preserving robust performance across varying perturbation budgets, confirming it as a reliable solution for adversarial scenarios.

\section{Conclusion} \label{sec:Conclusion}
In this work, we introduced \textbf{FedAPT}, a novel adversarial prompt tuning method designed to enhance the adversarial robustness of FPT under non-IID settings. \textbf{FedAPT} features a cross-modal \textbf{class-aware prompt generator}, which creates globally aligned visual prompts to bridge the \textit{class information gap} caused by data heterogeneity. The generator aggregates label information from multiple clients to construct a global \textit{beacon}, providing clients with global class knowledge. We also proposed a \textbf{cross-layer generator sharing} strategy, which ensures consistent prompts across model layers, further improving robustness and generalization. Extensive experiments demonstrate that FedAPT outperforms existing methods across 15 datasets (ImageNet,  10 cross-dataset and 4 cross-domain datasets), showing superior adversarial robustness and strong generalization capabilities in cross-domain and cross-dataset tasks. Our work offers a practical framework for secure fine-tuning of VLMs and sets a solid benchmark for safeguarding federated prompt tuning against adversarial attacks.

\bibliography{example_paper}
\bibliographystyle{icml2025}

\newpage
\appendix
\onecolumn

\end{document}